\PassOptionsToPackage{table}{xcolor}
\documentclass[10pt,twocolumn,letterpaper]{article}

\usepackage[pagenumbers]{cvpr} % To force page numbers, e.g. for an arXiv version

\usepackage{microtype}

\definecolor{cvprblue}{rgb}{0.21,0.49,0.74}
\usepackage[pagebackref,breaklinks,colorlinks,allcolors=cvprblue]{hyperref}

\usepackage{boldline,multirow,arydshln, xcolor}
\usepackage{graphicx}

\newcommand\pad{\vspace{-0.33cm}\\} % table padding
\usepackage{amssymb}% http://ctan.org/pkg/amssymb
\usepackage{pifont}% http://ctan.org/pkg/pifont
\newcommand{\cmark}{\ding{51}}%
\usepackage{wrapfig}
\usepackage{colortbl}
\usepackage{xcolor}
\usepackage{amsmath,amsfonts,bm, bbm}

\usepackage{tikz}

\def\eqref#1{equation~\ref{#1}}

\def\1{\bm{1}}

\DeclareMathAlphabet{\mathsfit}{\encodingdefault}{\sfdefault}{m}{sl}
\SetMathAlphabet{\mathsfit}{bold}{\encodingdefault}{\sfdefault}{bx}{n}

\def\gJ{{\mathcal{J}}}

\DeclareRobustCommand{\eg}{e.g.\@\xspace}

\newcommand{\ours}{\textup{\textsc{Atlas}}\xspace}
\newcommand{\dataset}{\textup{\textsc{OmniST}}\xspace}

\def\confName{CVPR}
\def\confYear{2027}

\title{Tracking Objects Through Assembly and Decomposition}

\author{Cheryl Lam \quad Yihong Sun \quad Bharath Hariharan\\
Cornell University
}
\begin{document}
\maketitle

% Teaser figure
\begin{figure*}
    \centering
    \includegraphics[width=1\linewidth]{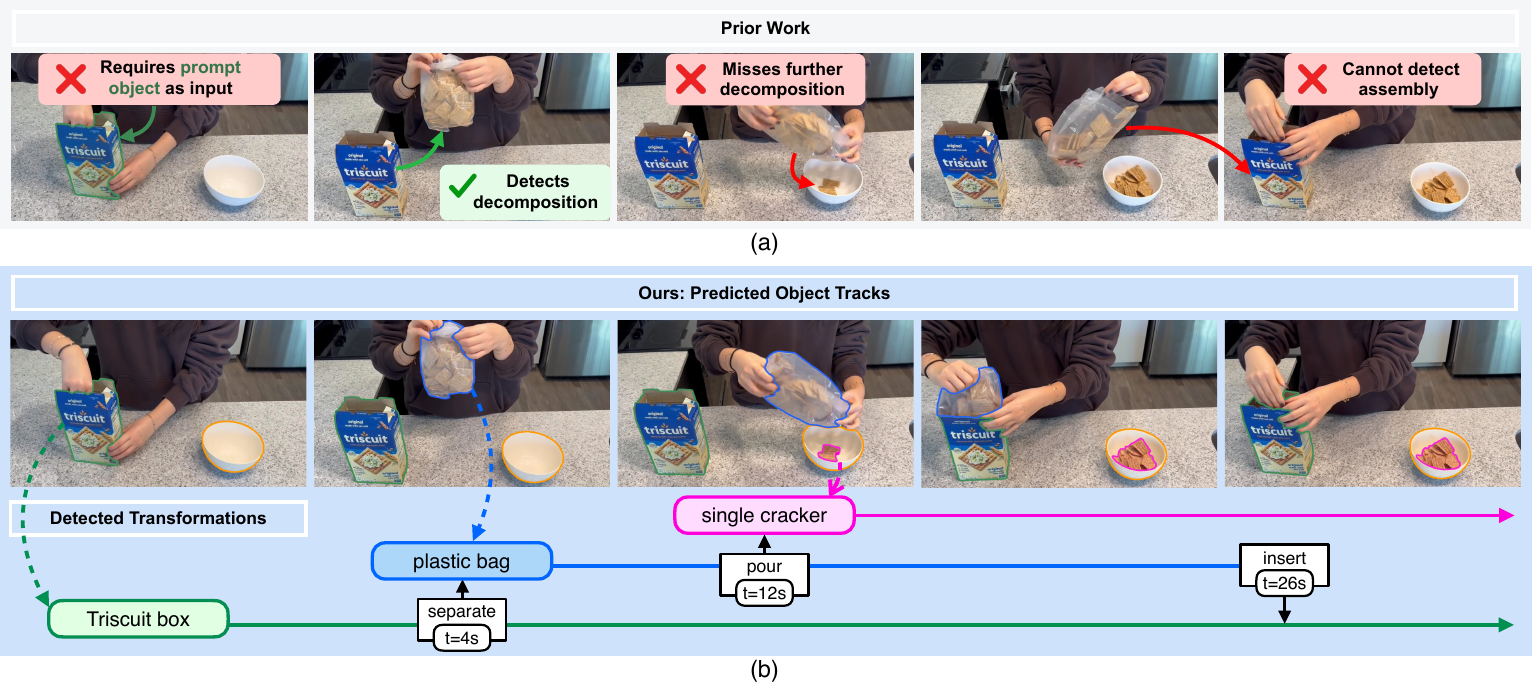}
    \caption{\textit{Top:} Prior methods~\cite{tubeletgraph, fluxgraph} require an initial object prompt and only detect its decomposition, but miss when objects resulting from that decomposition split further and cannot detect assembly events. \textit{Bottom:} Without a prompt, \ours recovers all object tracks and detects both decomposition and assembly. Object tracks are shown as overlaid colored masks, while detected transformations are visualized as a state graph linking the participating objects. For readability, only a subset of tracks is shown corresponding to objects participating in transformations and the bowl.}
    \label{fig:teaser}
\end{figure*}

% Main paper
\begin{abstract}
  % Abstract goes here.
  Tracking objects through transformations is essential for understanding real-world dynamics. Recent work has extended traditional object tracking to detect state changes. However, existing methods require an initial object prompt and detect only its decompositions, missing assembly events, which begin with multiple objects rather than one. We therefore introduce a prompt-free formulation of this task, where the goal is to recover all object tracks and detect both decomposition and assembly events given only a video. Our key insight is that assembly is decomposition in reverse. Building on this insight, we present \ours, a zero-shot system that constructs and jointly reasons over forward and reverse spatiotemporal partitions to recover object tracks and their transformations. We further introduce \dataset, a benchmark with dense annotations of objects and both transformation types. On \dataset, \ours recovers more than twice as many state changes as prior methods while remaining competitive on existing benchmarks that only evaluate decomposition. 
  Code, data, and additional results are available at \url{https://atlas-track-objects.github.io/}.
\end{abstract}
\section{Introduction}
\label{sec:intro}

Object transformations are everywhere: a chef slices a tomato; a child assembles a toy; a caterpillar becomes a chrysalis and later emerges as a butterfly. Understanding these changes is crucial, whether for reasoning about pre- and post-conditions of actions by embodied agents~\cite{liu2024blade,regneri2013grounding}, monitoring wildlife across developmental stages, or creating policy sub-goals from recorded demonstrations~\cite{pi_07}.

One approach is to ask a multimodal language model (MLLM) to identify transformations in a video~\cite{gpt4,team2024gemini}. Unfortunately, MLLMs provide only coarse descriptions of what occurred. Mounting evidence suggests that current MLLMs struggle to spatiotemporally ground temporal events in videos~\cite{flatpackbench, yu2026benchmarking}. This lack of grounding is problematic: e.g., it prevents embodied agents from tying transformations (``a boy building a toy") to the specific objects and context (the particular pieces and how they go together).

To ground transformations in the video, one must also track the objects involved. Recently, Sun et al.~\cite{tubeletgraph} extended traditional object tracking to detect state transformations as well. Their approach, TubeletGraph, takes a video and an initial target object mask as input, tracks the object through its transformations, and produces a state graph that records each transformation along with its timing, participating objects, and natural-language description.

However, this formulation has a critical flaw. Like traditional object tracking, the problem is formulated as tracking \emph{a single object} annotated in the first frame. This limits TubeletGraph~\cite{tubeletgraph} and other current systems~\cite{fluxgraph} to \emph{decomposition} events, where the tracked object fragments into independently trackable parts. When the reverse occurs—multiple objects \emph{merging} during assembly—these systems simply fail to detect the event, either silently ending the track or continuing it through the assembled object, resulting in incomplete and misleading output. Beyond assembly, current systems also miss any further decompositions of the fragments from an earlier decomposition since they only reason about the transformations of the original  prompted object. For example, in Figure~\ref{fig:teaser}, TubeletGraph detects the removal of the bag from the box, but misses the subsequent decomposition (removing crackers from the bag) and assembly (inserting the bag back into the box). 

To truly track \textit{any} transformation, we must relax the traditional tracking assumption of a single ``prompt'' object and instead account for all objects in the scene. Concretely, given only a video, the goal is to produce a spatiotemporal state graph recording every decomposition and assembly along with its timing, participating objects, and natural-language description. This increases task difficulty, as one must now search for all objects that may undergo a state change without any prompt to anchor the search.

To solve this, we present \textbf{\ours} (\textbf{A}ll-object \textbf{T}racking and \textbf{L}ocalization \textbf{A}cross \textbf{S}tate changes), a zero-shot framework for prompt-free tracking and understanding of object transformations. Our key insight is that \emph{assembly is decomposition in reverse}: a video of objects merging is, up to temporal direction, indistinguishable from one in which an object splits. Based on this insight, \ours constructs forward and reverse spatiotemporal partitions and jointly reasons across them using spatial relationships to recover state changes. \ours then uses a VLM to verify each candidate transformation, refine its timing, and describe the transformation. The name \ours also describes the output: a \emph{temporal atlas} in which, for any object at any time, one can retrieve its segmentation mask and the transformation subgraph leading to and following from its current state.

For evaluation, existing benchmarks such as VOST~\cite{vost} are also limited to decomposition events: assembly events, which start from multiple objects instead of one prompt object, are not even annotated in the data. We therefore create a new benchmark \dataset with annotated object tracks and both decomposition and assembly events. On \dataset, we find that \ours recovers \emph{more than twice as many} state changes as prior methods. At the same time, we maintain competitive performance on previous benchmarks that only evaluate decomposition~\cite{vost}.

In sum, the contributions of our work are:
\begin{enumerate}
    \item We introduce the task of tracking all objects and detecting their transformations given only a video.
    \item We propose \ours, a zero-shot framework that requires no initial prompt, constructs bidirectional spatiotemporal partitions, and models assembly as the temporal reverse of decomposition.
    \item We introduce \dataset, a benchmark with  decomposition and assembly events, and demonstrate state-of-the-art performance on \dataset while matching prior art on existing benchmarks.
\end{enumerate}
\section{Related Work}
\label{sec:related}

\textbf{Object Tracking and Transformation Understanding.}
Object tracking methods~\cite{vot2016} typically segment a target object specified by an initial mask and propagate it forward via appearance features~\cite{bhat2020learning,caelles2017one,hu2018videomatch,xmem,oh2019video,cutie}. SAM2~\cite{sam2} substantially advances this paradigm with a streaming memory-attention mechanism. Extensions such as SAM2Long~\cite{sam2long}, SAMURAI~\cite{samurai}, and DAM4SAM~\cite{dam4sam} address error accumulation, crowded scenes, and visual distractors. However, under dramatic visual changes, these trackers lose the target object. Datasets such as VOST~\cite{vost}, VSCOS~\cite{vscos}, DTTO~\cite{wu2024tracking}, and M$^3$-VOS~\cite{m3vos} benchmark this challenge. Re-VOS~\cite{m3vos} proposes combining forward and backward memory to improve robustness through transformations, while recent approaches improve SAM2 during state transformations, but require training~\cite{sam2tc}. However, these methods do not recover the transformations themselves. TubeletGraph~\cite{tubeletgraph} extends object tracking by introducing Track Any State: given an initial mask, it recovers objects missed after decomposition via a spatiotemporal partition and constructs a state graph through VLM reasoning. Our work directly builds on Track Any State, but removes the initial prompt requirement and extends the framework to handle assembly transformations.

Beyond prompt-based tracking, prompt-free and panoptic video segmentation methods~\cite{Li_2022_CVPR,cheng2023tracking, entitysam} track all objects in a scene simultaneously without a per-object prompt. However, these methods do not model transformations: splits and merges are treated as the appearance of new objects rather than state change events. Our method is the first to combine prompt-free all-object tracking with transformation understanding.

\vspace{2mm}
\noindent \textbf{Object State Change Recognition.}
A complementary line of work studies object state changes without maintaining persistent object tracks. HowToChange~\cite{oscs} and WhereToChange~\cite{spoc} train models to temporally and spatially localize three stages of a state change (initial, transitioning, and end) in instructional videos, using pseudo-labels from off-the-shelf vision-language systems. SPARTA~\cite{mandikal2025mash} demonstrates robotic manipulation capabilities built upon spatially-progressing state change segmentation. While these methods produce spatially grounded outputs at the progress level, they do not maintain consistent object identities across transformation boundaries. In contrast, our method produces persistent spatiotemporal tracks that can be queried after the fact for any object at any time.

\vspace{2mm}
\noindent \textbf{Vision and Language.} 
MLLMs such as GPT-4~\cite{gpt4} and Gemini~\cite{team2024gemini} can reason about visual and textual queries~\cite{lu2023chameleon,driess2023palm,wang2023voyager,gupta2023visual,suris2023vipergpt}, but lack pixel-level grounding and produce only coarse event-level descriptions. In our work, we use VLMs to verify, refine, and describe the state changes identified through spatiotemporal reasoning over tubelets. 
\section{Method}
\begin{figure*}
    \centering
    \includegraphics[width=1\linewidth]{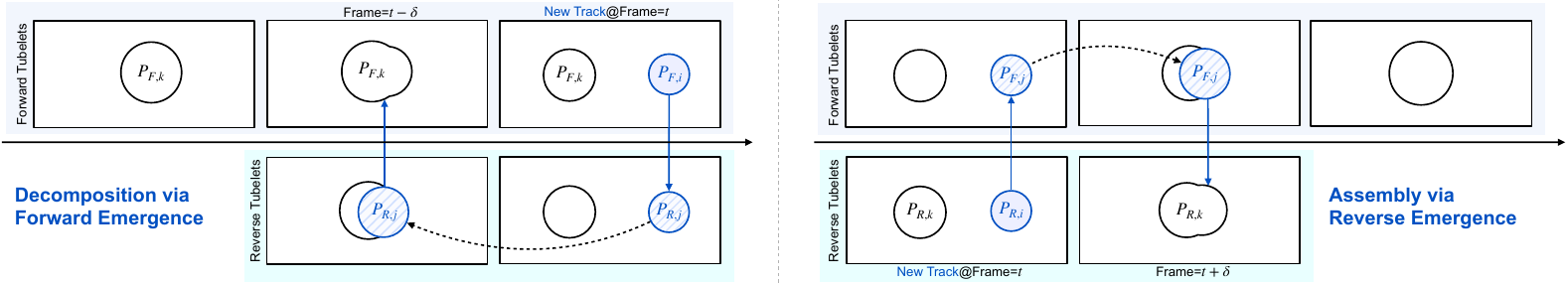}
    \caption{\textit{Left:} Tubelet $P_{F,i}$ emerges during forward tracking, matches $P_{R,j}$, and is identified as a decomposition product of $P_{F,k}$. \textit{Right:} Tubelet $P_{R,i}$ emerges during reverse tracking, matches $P_{F,j}$, and is identified as an object assembling into $P_{R,k}$. Solid blue arrows indicate matches across temporal partitions, while dotted black arrows trace the same tubelet through time.}
    \label{fig:method}
\end{figure*}

\subsection{Task Formulation}

We introduce a prompt-free formulation of Track Any State~\cite{tubeletgraph} that aims to recover all object tracks and identify their transformations (see Appendix~\ref{define_transformations} for an operative definition of transformations). 
Formally, given a video $V = \{I_t\}_{t=1}^T$, the desired output consists of:
\begin{enumerate}
    \item A collection of object tracks $\mathcal{T} = 
    \{\mathcal{T}^1, \ldots, \mathcal{T}^N\}$ where each
    track $\mathcal{T}^i$ is a sequence of
    masks $\{\mathcal{M}_t^i\}$; and 
    \item A collection of transformations / state changes $\mathcal{S}=
    \{s\}$ where $s = (t,
    \mathcal{T}_{\text{pre}},
    \mathcal{T}_{\text{post}},
    D)$ with $t$ denoting the frame where the change
    happened,
    $\mathcal{T}_{\text{pre}},
    \mathcal{T}_{\text{post}}
    \subseteq \mathcal{T}$ the sets of tracks involved
    before and after the change, and $D$ a natural
    language description.
\end{enumerate}
Figure~\ref{fig:teaser} (bottom) shows an example output with tracks visualized as overlaid colored masks and state changes shown as a graph.
The original formulation of TubeletGraph~\cite{tubeletgraph} is a special case where an initial object mask is an added input, and the set of output object tracks and state changes are restricted to the subset involving the prompt object.

\subsection{Preliminaries}

We build on the observation in TubeletGraph~\cite{tubeletgraph} that the decomposition of an object results in the appearance of new tracks. For example, in Figure ~\ref{fig:teaser} (top), when the  bag is removed from the box (the prompt object), the tracker~\cite{sam2} follows the box, leaving the bag untracked. TubeletGraph initializes a new track for this untracked region and reasons about its relation to the prompt object to identify the transformation.  However, TubeletGraph cannot detect assembly, as assembly does not trigger the appearance of new tracks during forward tracking.

Since assembly is decomposition in reverse, tracking backward allows assembly to be detected through the emergence of new tracks in the reverse direction. We therefore build spatiotemporal partitions~\cite{tubeletgraph} in both the forward and the backward directions. We then consider tracks that \emph{newly emerge} in the middle of the video in either direction, and use \emph{both forward and reverse partitions} to reason about whether these tracks are the signatures of decomposition or assembly events. 

We now describe each step of this pipeline in detail.

\subsection{Building Bidirectional Spatiotemporal Partitions} \label{subsec:reasoning}

We construct the forward spatiotemporal partition of the video following TubeletGraph~\cite{tubeletgraph}: we segment every frame with CropFormer~\cite{cropformer}, propagate every entity in the first frame forward with SAM2~\cite{sam2}, and at each subsequent frame, initialize a new track for any entity whose area is covered by less than $\tau_{\text{coverage}}$ of existing tracks. During propagation, multiple tracks may collapse onto the same object (e.g., during assembly). To maintain a set of spatially distinct entities, if two tracks that have each been tracked for more than 5 frames share an IoU greater than $\tau_\text{dup}$, we terminate the track with lower confidence. We repeat this process in reverse to obtain the reverse partition (see the supplementary for video visualizations of both partitions). We represent the forward and reverse partitions as pools of tubelets, $\mathcal{P}_F = \{P_{F,i}\}$ and $\mathcal{P}_R = \{P_{R,i}\}$, where each tubelet $P_i = \{e_t^i\}$ consists of masks $e_t^i$ for entity $i$ at frame $t$.

\subsection{Detecting State Changes} 

Having obtained forward and reverse partitions of the video, we now focus on detecting any transformations that occur. A tubelet that newly emerges in either temporal direction may correspond either to an object produced through a transformation (decomposition in forward or assembly in reverse) or to a newly introduced object entering the field of view. To distinguish these cases, we assume that tubelets arising from transformations spatially originate from previously tracked objects while newly introduced objects do not. We use this assumption to define two predicates for detecting state changes. Throughout, we say that a tubelet $P_A=\{e_t^A\}$ \emph{matches} $P_B=\{e_t^B\}$ at frame $t$ when $e_t^A$ is contained within $e_t^B$ and the two masks have sufficient IoU:
\[
\frac{|e_t^{A}\cap e_t^{B}|}{|e_t^{A}|} > \tau_{\text{contain}},
\qquad
\frac{|e_t^{A}\cap e_t^{B}|}{|e_t^{A}\cup e_t^{B}|} > \tau_{\text{IoU}}.
\]

\textbf{Decomposition via Forward Emergence.}
Suppose a tubelet $P_{F,i}$ newly emerges at frame $t>1$ in the forward tracking direction, as shown in Figure~\ref{fig:method} (left). To determine whether this tubelet corresponds to a transformation product or a newly introduced object, we look at all reverse tubelets at frame $t$. If $P_{F,i}$ matches a reverse tubelet $P_{R,j}$, we search backward in time for up to $\omega$ seconds. If, at an earlier frame $t-\delta$ within this window, $P_{R,j}$ matches a previously tracked tubelet $P_{F,k}$, we identify a candidate decomposition in which $P_{F,i}$ originated from $P_{F,k}$. We mark $P_{F,i}$ and $P_{R,j}$ as duplicates since they correspond to the same post-transformation object and merge them during final tubelet construction. We record the state change as $(t,\{P_{F,k}\},\{P_{F,k},P_{F,i}\})$.

\textbf{Assembly via Reverse Emergence.}
Similarly, suppose a tubelet $P_{R,i}$ newly emerges at frame $t<T$ in the reverse tracking direction, as shown in Figure~\ref{fig:method} (right). We compare it with all forward tubelets at frame $t$. If $P_{R,i}$ matches $P_{F,j}$, we search forward in time for up to $\omega$ seconds. If, at a later frame $t+\delta$ within this window, $P_{F,j}$ matches $P_{R,k}$, we identify a candidate assembly in which $P_{R,i}$ merges into $P_{R,k}$. We mark $P_{R,i}$ and $P_{F,j}$ as duplicates since they correspond to the same pre-transformation object and merge them during final tubelet construction. We record the state change as $(t,\{P_{R,i},P_{R,k}\},\{P_{R,k}\})$. This yields a set of candidate transformations.

\subsection{Transformation Verification and Realignment}
\begin{figure*}
    \centering
    \includegraphics[width=1\linewidth]{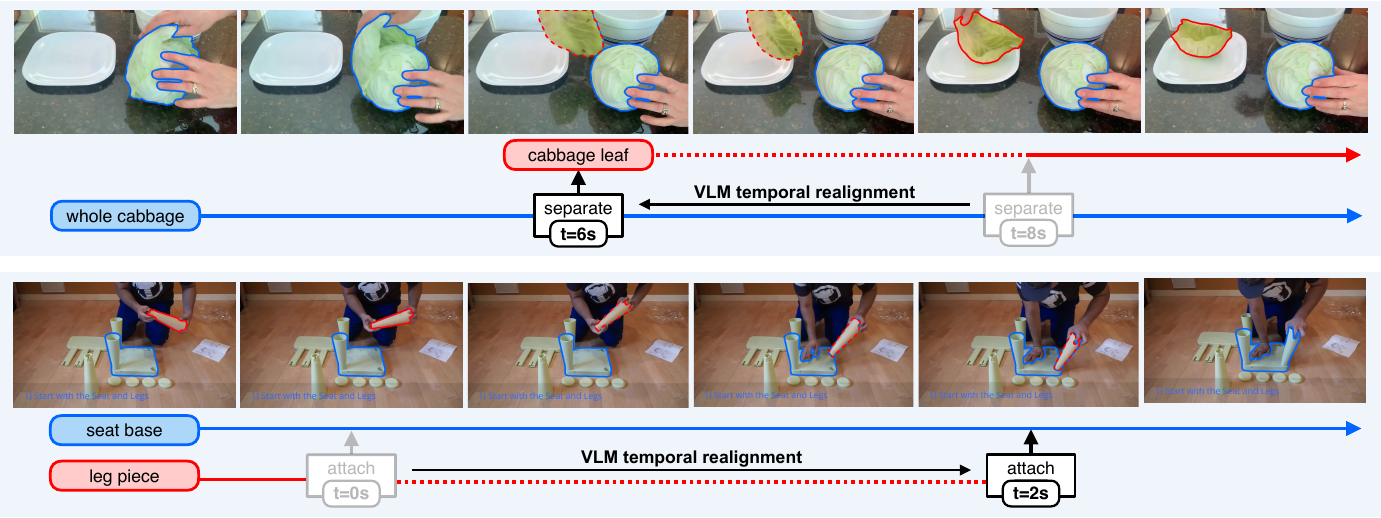}
    \caption{A VLM temporally realigns a decomposition event detected late (top) and an assembly event detected early (bottom).}
    \label{fig:realign}
\end{figure*}
At this point, the candidate transformations have two potential issues. First, our detected state changes may not represent actual state changes due to coincidental spatial overlap. Second, the detected frame $t$ may differ from the frame $t_{\text{true}}$ at which the transformation actually occurs. In a decomposition, SAM2~\cite{sam2} often keeps a single mask over the separating objects for several frames, so $P_{F,i}$ emerges at some $t > t_{\text{true}}$. In reverse tracking, SAM2 keeps a single mask over the merged objects past the point where they separate in reverse time, so $P_{R,i}$ emerges at some $t < t_{\text{true}}$. Thus, decompositions are often detected late and assemblies early.

We address both issues simultaneously with a VLM. Although a VLM cannot spatially ground transformations, we leverage it for semantic and temporal reasoning about candidate events. For each candidate transformation, we prompt the VLM with full frames sampled uniformly from the video. We then provide frames sampled around the predicted event, each cropped to the union of the participating objects' bounding boxes with their contours overlaid. The full frames give context for the whole video, and the crops show the objects closely enough to verify the transformation and locate it in time. Based on this observation that decompositions are predicted late and assembly events are predicted early, we sample frames asymmetrically. For a suspected decomposition at $t$, we sample frames from $[t - \omega_{\text{long}}, t +  \omega_{\text{short}}]$ where $\omega_{\text{short}} < \omega_{\text{long}}$. For a suspected assembly at $t$, we sample frames from $[t - \omega_{\text{short}}, t +  \omega_{\text{long}}]$. We ask the VLM to determine whether a transformation actually occurred. If the VLM confirms a transformation, it outputs the frame at which separation first becomes visually apparent for decompositions or merging becomes visually complete for assembly events. The returned frame index is used as the final event time (Figure ~\ref{fig:realign}). The VLM also names the action and the objects involved in the pre- and post-transformation states, completing our state changes $\mathcal{S}$.

\subsection{Producing a Unified Spatiotemporal Partition}
Once we have detected the state transformations, we proceed to produce a final, unified set of tubelets that form our spatiotemporal partition.

We begin by identifying duplicate tubelets across the forward and reverse partitions $\mathcal{P}_F$ and $\mathcal{P}_R$. For every pair $(P_{F,i}, P_{R, j}) \in \mathcal{P}_F \times \mathcal{P}_R$, we compute the video-level mean IoU over their temporal overlap. We then perform Hungarian matching~\cite{hungarian} between forward and reverse tubelets, retaining matched pairs exceeding $\tau_{\text{merge}}$, along with the matches identified during transformation detection. For each matched pair, we take the union of their timestamps and masks, producing a single tubelet. We initialize our final set of tracks $\mathcal{T}$ with these unified tubelets and update the pre- and post-tracks of each detected transformation to point to this set. We also add tracks involved in a transformation without a duplicate in the other direction to $\mathcal{T}$.

Finally, we clean up the tracks $\mathcal{T}$ both spatially and temporally. For each state change, guided by the VLM's temporal alignment, we trim the tracks accordingly: for a decomposition predicted at frame $t$, we enforce that the emergent object's track begins at $t$; for an assembly predicted at frame $t$, we enforce that the emergent track ends at $t$. To prevent mask overlap, we subtract the emergent object's mask from the source tubelet's mask at each frame. To ensure complete spatiotemporal coverage, we add to the final pool any tubelet from either tracking direction that has neither been matched as a duplicate nor associated with a state change if less than $\tau_{\text{area}}$ of its area is already covered by tubelets in the pool. The final pool is a full spatiotemporal partition of the video. 

\section{Dataset}

Previous datasets such as VOST-TAS~\cite{tubeletgraph}, curated from the VOST validation set~\cite{vost}, only contain annotations for decomposition events and do not capture assembly. We therefore introduce a more comprehensive dataset  with \emph{both decomposition and assembly events}.
We source our videos from (1) recordings by the authors, (2) YouTube tutorials retrieved via action-object queries (e.g., ``how to replace Swiffer mop pad'', ``how to mount a car tire''), and (3) existing benchmarks including VOST~\cite{vost}, VSCOS~\cite{vscos}, M$^3$-VOS~\cite{m3vos}, and IKEA Video Manuals~\cite{liu2024ikea}.

In each video, we annotate decomposition and assembly events when all participating objects can be clearly identified before and after the transformation. Each transformation is a 4-tuple $\tau_i = (T_i, v_i, \mathcal{O}_i^{\text{pre}}, \mathcal{O}_i^{\text{post}})$, where $T_i = [t_i^s,t_i^e]$ is the temporal span and $v_i$ is a free-text description. The sets $\mathcal{O}_i^q = \{(o,d_{i,o}^q) \mid o \in P_i^q\}$, for $q \in \{\text{pre},\text{post}\}$, specify the participating objects $P_i^q$ and their names $d_{i,o}^q$ before and after the transformation. Each object $o$ is annotated with segmentation masks $\{m_o^t\}_{t \in \mathcal{T}_o}$ over all frames $\mathcal{T}_o$ where it is visible.

Table~\ref{tab:dataset} compares \dataset with VOST-TAS. More details about the dataset collection and qualitative examples are presented in Appendix~\ref{dataset_details}.

\begin{table}[t]
\centering
\begin{tabular}{lcc}
\toprule
& \textbf{VOST-TAS} & \textbf{\dataset (Ours)}  \\
\midrule
Videos              & 57     & 97 \\
Ann fps.            & 5      & 15 \\
Objects             & 174    & 352 \\
Object masks        & 350    & 54,369 \\
\midrule
Transformations         & 108       & 302 \\
\quad Assembly          & --    & 144 \\
\quad Decomposition     & 108       & 158 \\

\bottomrule
\end{tabular}
\caption{Comparison of VOST-TAS~\cite{tubeletgraph} and \dataset. VOST-TAS annotates only decomposition events with object masks at select frames, while \dataset annotates both decomposition and assembly events with dense per-frame masks of participating objects.}
\label{tab:dataset}
\end{table}

\begin{table*}[t]
\centering
\setlength{\tabcolsep}{4pt}
\renewcommand{\arraystretch}{0.95}

\begin{tabular*}{\textwidth}{@{\extracolsep{\fill}}l c cc cc cc cc@{}}
\toprule

&
& \multicolumn{8}{c}{\textbf{Transformation Detection}} \\

\cmidrule(lr){3-10}

\textbf{Method}
& \textbf{Tracking}
& \multicolumn{2}{c}{Temp. Loc.}
& \multicolumn{2}{c}{Sem. Acc.}
& \multicolumn{2}{c}{Spatiotemp.}
& \multicolumn{2}{c}{Overall} \\

\cmidrule(lr){2-2}
\cmidrule(lr){3-4}
\cmidrule(lr){5-6}
\cmidrule(lr){7-8}
\cmidrule(lr){9-10}

& $\gJ_{\text{tubelet}}$
& $\mathcal{T}_P$ & $\mathcal{T}_R$
& $\mathcal{S}_V$ & $\mathcal{S}_O$
& $\mathcal{H}_{ST,P}$ & $\mathcal{H}_{ST}$
& $\mathcal{H}_P$ & $\mathcal{H}$ \\

\midrule

GPT-4.1 + GDINO
& 52.5 & \textbf{80.1} & \textbf{65.2} & \textbf{90.4} & 95.8 & 9.8 & 7.9 & 7.7 & 6.3 \\

GPT-4.1 + SAM3
& 53.6 & \textbf{80.1} & \textbf{65.2} & \textbf{90.4} & \underline{96.5} & 19.1 & 15.6 & 13.8 & 11.3 \\

\midrule

TubeletGraph
& \underline{57.3}
& 37.7 & 26.5
& 80.0 & 91.0
& 22.6 & 15.9
& 16.5 & 11.6 \\

FluxGraph
& 48.0
& 39.4 & 16.6
& 82.0 & 88.9
& 23.6 & 9.9
& 16.5 & 7.0 \\

TubeletGraph-R
& 52.7
& 30.2 & 18.9
& 52.6 & 64.0
& 13.2 & 8.3
& 5.8 & 3.6 \\

TubeletGraph-Bidir
& 56.0
& 34.2 & 45.4
& 67.9 & 83.3
& 18.5 & 24.5
& 12.2 & 16.2 \\

\quad + Verif. \& Realign.
& 55.9
& 64.0 & 50.7
& 87.6 & 95.6
& \underline{35.1} & \underline{27.8}
& \underline{29.3} & \underline{23.2} \\

\midrule

\textbf{\ours}
& \textbf{68.7}
& \underline{77.4} & \underline{57.9}
& \underline{90.3} & \textbf{96.9}
& \textbf{57.1} & \textbf{42.7}
& \textbf{48.2} & \textbf{36.1} \\

\bottomrule
\end{tabular*}

\caption{Comparison of \ours with baselines on object tracking and transformation detection on \dataset. ``+ Verif.\ \& Realign.'' applies our VLM verification and temporal realignment stage to TubeletGraph-Bidir. \ours significantly outperforms all baselines on $\gJ_{\text{tubelet}}$, $\mathcal{H}_{ST,P}$, $\mathcal{H}_{ST}$, $\mathcal{H}_P$, and $\mathcal{H}$. Best performance is bolded and second bests are underlined. We report 95\% confidence intervals in Appendix~\ref{additional_results}. }
\label{tab:main}

\end{table*}

\section{Experiments}
\label{sec:exp}

% Experiments goes here.

\subsection{Experimental Setup}

\textbf{Datasets.}
We evaluate object tracking and transformation detection on \dataset.
Following TubeletGraph~\cite{tubeletgraph}, we additionally evaluate object tracking on VOST~\cite{vost}, VSCOS~\cite{vscos}, and M$^3$-VOS~\cite{m3vos}, and transformation detection on VOST-TAS~\cite{tubeletgraph}.
These benchmarks provide object masks in the first frame as prompts, which we incorporate into \ours for comparison with prompt-based methods.
For M$^3$-VOS, we exclude 20 videos with annotations that differ from our treatment of assembly, and evaluate all methods on the same remaining videos. Further details on evaluation on existing benchmarks are provided in Appendix~\ref{prompt_based_eval}.

\textbf{Metrics.} For object tracking on existing benchmarks, which contain only decomposition transformations, we follow VOST~\cite{vost} and report Jaccard $\gJ$ over the entire video and $\gJ_{\mathrm{tr}}$ over the last 25\% of frames, computed on the union of the masks of the prompt object and its resulting fragments. However, distinct objects combine during assembly, requiring tubelets to be evaluated individually. For object tracking on \dataset, we therefore construct a new tracking metric analogous to $\gJ$ that evaluates individual tubelets. Since object identities may be ambiguous during transformations, we exclude frames within annotated transformation intervals and split ground-truth tracks into temporally stable tubelets at transformation boundaries. We similarly split predicted tracks at their predicted transformation boundaries. We then apply Hungarian matching~\cite{hungarian} between predicted and ground-truth tubelets using mean IoU over all frames of the ground-truth tubelet. We report $\gJ_\text{tubelet}$, the average matched tubelet IoU over objects participating in transformations, with unmatched ground-truth tubelets assigned zero.

To evaluate transformation detection on VOST-TAS and \dataset, we follow the evaluation protocol of TubeletGraph~\cite{tubeletgraph}, additionally reporting spatiotemporal and overall precision for \dataset:
\begin{enumerate}
\item \textbf{Temporal localization precision and recall} ($\mathcal{T}_P$, $\mathcal{T}_R$) measure whether predicted transformations fall within ground-truth intervals, extended by $0.5$\,s on each side. $\mathcal{T}_P$ is the fraction of detections correctly localized, while $\mathcal{T}_R$ is the fraction of ground-truth transformations correctly localized.
\item For temporally localized transformations, we report \textbf{semantic accuracy} for predicted action verbs ($\mathcal{S}_V$) and predicted object descriptions ($\mathcal{S}_O$) for objects that are matched with IoU $> 0.5$. Verb and object correctness are judged by GPT-4.1~\cite{gpt4}.
\item \textbf{Spatiotemporal precision and recall} ($\mathcal{H}_{ST,P}$, $\mathcal{H}_{ST}$) require both successful temporal localization and participating objects to be matched with IoU $>0.5$ at the extended annotated boundaries.
\item \textbf{Overall precision and recall} ($\mathcal{H}_P$, $\mathcal{H}$) require successful
spatiotemporal localization and correct verb and object descriptions.
\end{enumerate}

\textbf{Implementation Details.}
\ours uses the same backbone models as TubeletGraph: CropFormer-Hornet3X~\cite{cropformer} for entity segmentation, SAM2.1-L~\cite{sam2} for tracking, and GPT-4.1~\cite{gpt4} (temperature $0$) for state graph construction. 
Thresholds are selected once on a separate set of videos; see Appendix~\ref{robustness} for analysis of robustness to these choices.

\begin{table*}[t]
\centering
\resizebox{\textwidth}{!}{%
\begin{tabular}{lc@{\hspace{3pt}}c c@{\hspace{6pt}}c@{\hspace{6pt}}c@{\hspace{6pt}}c@{\hspace{6pt}}c@{\hspace{6pt}}c c@{\hspace{6pt}}c@{\hspace{6pt}}c@{\hspace{6pt}}c@{\hspace{6pt}}c@{\hspace{6pt}}c}
\toprule
& \multicolumn{2}{c}{} &
\multicolumn{6}{c}{\textbf{Tracking}} &
\multicolumn{6}{c}{\textbf{Transformation Detection}} \\
\pad \cline{4-15} \pad
\textbf{Method} &
\multicolumn{2}{c}{\textbf{Detects}} &
\multicolumn{2}{c}{VOST} &
\multicolumn{2}{c}{VSCOS} &
\multicolumn{2}{c}{M$^3$-VOS} &
\multicolumn{2}{c}{Sem. Acc.} &
\multicolumn{2}{c}{Temp. Loc.} &
\multicolumn{2}{c}{Overall} \\
\pad \cline{2-15} \pad
&
Decomp. &
Assem. &
$\mathcal{J}$ & $\mathcal{J}_{tr}$ &
$\mathcal{J}$ & $\mathcal{J}_{tr}$ &
$\mathcal{J}$ & $\mathcal{J}_{tr}$ &
$\mathcal{S}_V$ & $\mathcal{S}_O$ &
$\mathcal{T}_P$ & $\mathcal{T}_R$ &
$\mathcal{H}_{ST}$ & $\mathcal{H}$ \\

\midrule
Cutie~\cite{cutie}          & - & - &41.1 &25.5 &70.9 &67.1 &\textbf{76.0} &\textbf{66.6} & - & - & - & - & - & - \\
ReVOS~\cite{m3vos}          & - & - &41.0 &25.3 &-    &-    &- &- & - & - & - & - & - & - \\
SAM2~\cite{sam2}            & - & - &46.1 &29.4 &72.5 &67.1 &72.4 &61.3 & - & - & - & - & - & - \\
SAM2Long~\cite{sam2long}    & - & - &46.4 &29.1 &73.0 &68.6 &70.8 &59.0 & - & - & - & - & - & - \\
SAM2.1~\cite{sam2}          & - & - &48.4 &32.4 &72.0 &66.9 &72.8 &61.1 & - & - & - & - & - & - \\
DAM4SAM~\cite{dam4sam}      & - & - &48.8 &33.6 &71.3 &66.0 &72.0 &61.2 & - & - & - & - & - & - \\
SAMURAI~\cite{samurai}      & - & - &49.8 &34.0 &71.8 &66.9 &73.5 &62.7 & - & - & - & - & - & - \\
\midrule
TubeletGraph~\cite{tubeletgraph}
& \cmark & - &
50.9 &36.7 &
\textbf{75.9} & \textbf{72.2} &
75.8 &66.4 &
81.8 &  72.3 &
43.1 & \textbf{20.4} &
\textbf{12.0} &6.5 \\

FluxGraph~\cite{fluxgraph}
& \cmark & - &
\textbf{52.8} & \textbf{36.8} &
75.3 &70.7 &
75.3 &65.9 &
77.3 & 82.9 &
36.7 &\textbf{20.4} &
\textbf{12.0} &9.3 \\

\ours 
& \cmark & \cmark &
50.2 & 34.6 &
74.7 & 70.7 &
73.5 & 62.9 &
\textbf{89.5} &\textbf{97.3} &
\textbf{51.4} & 17.6 &
11.1 &\textbf{10.2} \\

\bottomrule
\end{tabular}
}
\caption{Quantitative results for object tracking on VOST~\cite{vost}, VSCOS~\cite{vscos} and M\textsuperscript{3}-VOS~\cite{m3vos} and transformation detection on VOST-TAS~\cite{tubeletgraph}.}
\label{tab:benchmark}
\end{table*}

\subsection{Baselines} On \dataset, we compare \ours with VLM-based baselines and prompt-free adaptations of TubeletGraph~\cite{tubeletgraph}:

\begin{itemize}
    \item \textbf{GPT-4.1 + GDINO} uses GPT-4.1~\cite{gpt4} to predict transformation events directly from sampled video frames, then grounds the referenced objects with Grounding DINO~\cite{liu2023grounding} and tracks them with SAM2~\cite{sam2}.
    \item \textbf{GPT-4.1 + SAM3} uses the same GPT-4.1 predictions, but grounds and tracks the referenced objects with SAM3~\cite{carion2025sam3segmentconcepts}.
    \item \textbf{TubeletGraph}~\cite{tubeletgraph} runs the original TubeletGraph pipeline initialized with all CropFormer entities~\cite{cropformer} in the first frame.
    \item \textbf{FluxGraph}~\cite{fluxgraph} is an efficient variant of TubeletGraph.
    \item \textbf{TubeletGraph-R} runs TubeletGraph on the reversed video, so that assemblies appear as decompositions and can be detected.
    \item \textbf{TubeletGraph-Bidir} runs TubeletGraph independently in both directions to detect decompositions and assemblies, unlike \ours, which jointly reasons over both partitions (Figure~\ref{fig:method}). It then matches forward and reverse tracks via Hungarian matching~\cite{hungarian} and merges matched pairs when their average per-frame IoU exceeds $0.75$. We also apply our verification and realignment stage to it (\textbf{+ Verif.\ \& Realign.}), which serves as an ablation of \ours without joint reasoning.
\end{itemize}

\subsection{Results on \textbf{\dataset}}
Table~\ref{tab:main} compares \ours with all baselines on \dataset. \ours achieves the highest object tracking score of $\gJ_\text{tubelet}=68.7$, outperforming TubeletGraph~\cite{tubeletgraph} by $11.4$ points.

For transformation detection, \ours achieves the highest spatiotemporal and overall metrics. In particular, \ours achieves over $2\times$ the overall recall ($\mathcal{H}$) of TubeletGraph-Bidir, improving from $16.2$ to $36.1$, while achieving $4\times$ the overall precision ($\mathcal{H}_P$), improving from $12.2$ to $48.2$. A breakdown by transformation type in Appendix~\ref{additional_results} shows that \ours improves overall recall over TubeletGraph-Bidir from $9.0$ to $34.7$ on assembly events and from $22.8$ to $37.3$ on decomposition events. Even applying our VLM stage to TubeletGraph-Bidir (+ Verif.\ \& Realign.) closes only part of this gap, reaching $\mathcal{H}=23.2$ and $\mathcal{H}_P=29.3$. The remaining gap reflects the contribution of jointly reasoning over the two partitions, suggesting that running TubeletGraph independently in both directions is not enough. 

To understand where these gains come from, we analyze the transformations that \ours recovers but TubeletGraph-Bidir misses. First, as discussed, TubeletGraph misses decompositions of fragments produced earlier in the video (\eg, crackers removed from a bag that was first taken out of its box). \ours recovers $8$ such decompositions in \dataset. Second, TubeletGraph accepts a new tubelet as a transformation product only if it satisfies both a spatial proximity prior and a semantic prior which requires high FC-CLIP~\cite{fcclip} feature similarity between the new tubelet and the prompt object and often fails for assembly. In $13$ transformations recovered by \ours, TubeletGraph's spatial proximity prior was satisfied but not its semantic prior (\eg, an SD card inserted into a camera, a pair of sunglasses inserted into an eyeglass case, a cleaning cloth removed from an eyeglass case). 

On the other hand, VLM-based baselines achieve the highest temporal localization and high semantic accuracy, demonstrating that a VLM can reliably identify what is happening and when. For temporal localization, VLM-based baselines outperform \ours, which misses some transformations that do not result in new tubelets. However, GPT-4.1 + SAM3 yields only about a third of \ours's spatiotemporal recall and overall recall ($\mathcal{H}_{ST} = 15.6$, $\mathcal{H} = 11.3$), indicating that it struggles to localize the objects involved in transformations.

\begin{figure*}
    \centering
    \includegraphics[width=1\linewidth]{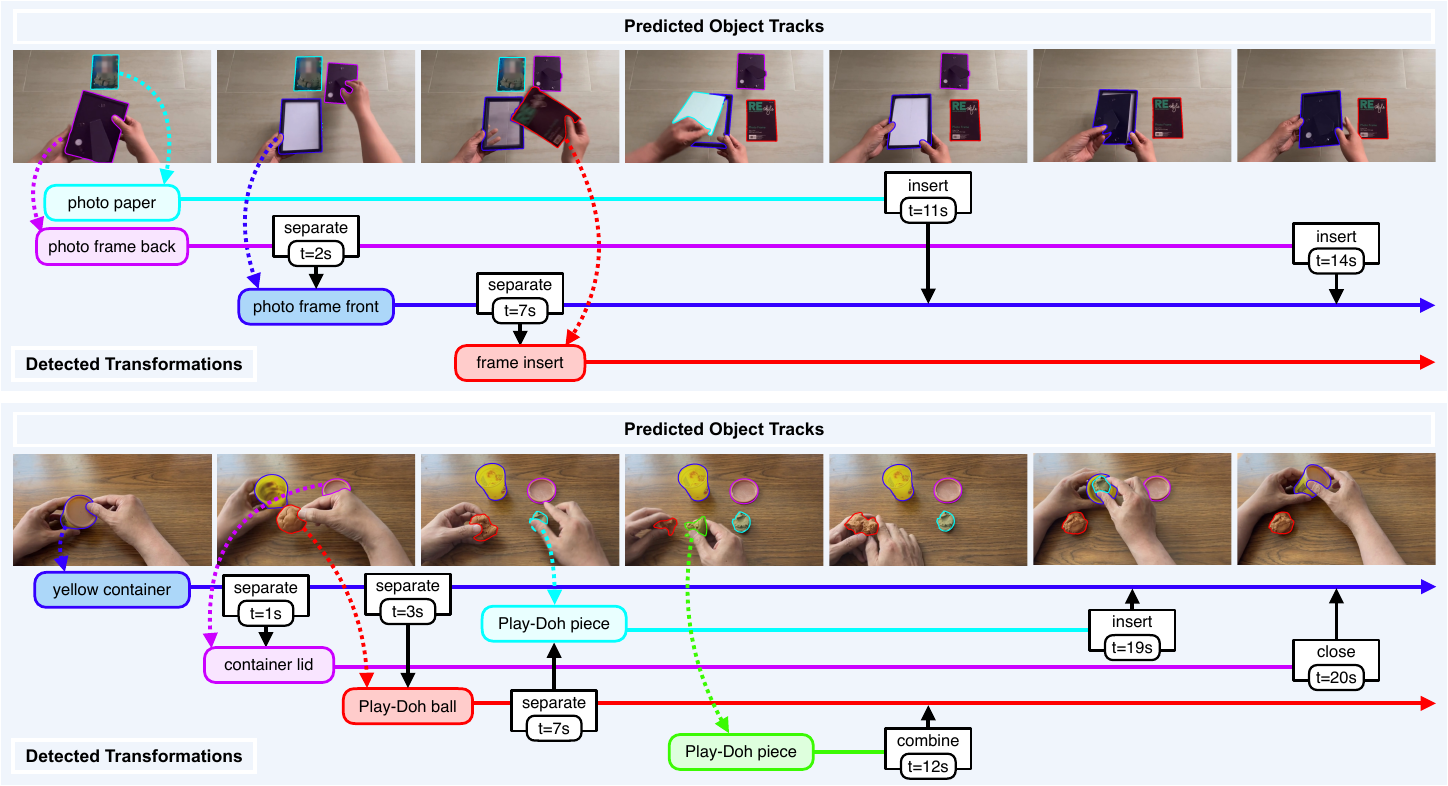}
    \caption{Qualitative results on \dataset. While \ours recovers tracks for all objects in a video, only the tracks of objects involved in the detected transformations are shown for readability. See the supplementary for additional qualitative results and comparisons with baselines.}
    \label{fig:qual}
\end{figure*}

\subsection{Results on Benchmarks Without Assembly} 

Even though VOST~\cite{vost}, VSCOS~\cite{vscos}, and M$^3$-VOS~\cite{m3vos} do not capture assembly transformations, Table~\ref{tab:benchmark} shows that \ours maintains tracking performance comparable to prompt-based methods on these benchmarks while being the only method that recovers both decomposition and assembly events. 

For transformation detection on VOST-TAS~\cite{tubeletgraph}, \ours has the highest semantic verb and object accuracy, improving over TubeletGraph from $\mathcal{S}_V=81.8$ to $89.5$ and from $\mathcal{S}_O=72.3$ to $97.3$. Since both methods use GPT-4.1~\cite{gpt4}, we attribute these gains to providing the VLM with additional temporal and spatial information through frames sampled throughout the video and around each emergence, unlike TubeletGraph, which reasons from only the initial frame and the emergence frame. \ours also improves temporal precision from $43.1$ to $51.4$, as it uses GPT-4.1 to reject false positives, while temporal recall decreases slightly from $20.4$ to $17.6$, as verification may also filter valid transformations.

\subsection{Ablations} 

We ablate the two uses of the VLM in \ours (Table~\ref{tab:ablate_vlm}). Without verification, precision decreases substantially: $\mathcal{T}_P$ falls from $77.4$ to $64.6$ and $\mathcal{H}_P$ from $48.2$ to $34.7$, while $\mathcal{T}_R$ rises slightly from 57.9 to 62.9. Verification trades a small amount of recall for the ability to filter out false positive transformations. Without realignment, both temporal precision and recall drop, reducing $\mathcal{H}$ from $36.1$ to $25.2$. Finally, disabling verification and realignment produces the largest degradation, reducing $\mathcal{H}$ to $22.8$, demonstrating that VLM reasoning is essential for accurate transformation detection.

\subsection{Qualitative Results}

Figure~\ref{fig:qual} shows qualitative results on two sequences with both decomposition and assembly events. The top example, replacing a photograph in a picture frame, illustrates accurate recovery of all objects and their transformations. The bottom example, manipulating Play-Doh, shows \ours on a longer sequence with more transformations.

\begin{table}[t]
\centering
\small
\setlength{\tabcolsep}{1.9pt}
\renewcommand{\arraystretch}{0.95}
\resizebox{\columnwidth}{!}{%
\begin{tabular}{lcccccc}
\toprule
\textbf{Variant}
& $\mathcal{T}_P$ & $\mathcal{T}_R$
& $\mathcal{H}_{ST,P}$ & $\mathcal{H}_{ST}$
& $\mathcal{H}_P$ & $\mathcal{H}$ \\
\midrule
\textbf{Full Model}
& \textbf{77.4} & 57.9
& \textbf{57.1} & 42.7
& \textbf{48.2} & \textbf{36.1} \\
w/o Verification
& 64.6 & \textbf{62.9}
& 44.9 & \textbf{43.7}
& 34.7 & 33.8 \\
w/o Realignment
& 63.1 & 41.4
& 48.5 & 31.8
& 38.4 & 25.2 \\
w/o Verification \& Realignment
& 48.5 & 47.4
& 31.9 & 31.1
& 23.4 & 22.8 \\
\bottomrule
\end{tabular}%
}
\caption{Ablation of VLM-based verification and realignment.}
\label{tab:ablate_vlm}
\end{table}

\section{Conclusion}
\label{sec:conclusion}

In this work, we introduce a prompt-free formulation of Track Any State, where the goal is to track all objects and detect their transformations. Since assembly is decomposition in reverse, we present \ours, a zero-shot system that jointly reasons over forward and reverse spatiotemporal partitions to recover object tracks and construct a state graph. We also introduce \dataset, a benchmark with dense annotations of objects and both transformation types. On \dataset, \ours more than doubles the overall recall of prior methods, while remaining competitive on benchmarks that annotate only decomposition.

\section*{LLM Usage Statement} 
We used generative AI tools (Claude Code and Codex) to assist with code implementation, improve the clarity and readability of the manuscript, and obtain reviewer-style feedback. These tools were not used for research ideas, dataset creation, experimental design, or evaluation and interpretation of the results. The authors take full responsibility for the final work.

% References
{
    \small
    \bibliographystyle{ieeenat_fullname}
    \bibliography{main}
}

% Supplementary material
% Remove or comment out for a main-paper-only submission.
\appendix
% \newpage
% \section{Appendix}    

\clearpage
\setcounter{page}{1}
\maketitlesupplementary
\section{Appendix}

% We provide a supplementary webpage, \texttt{index.html}, with visualizations of the forward and reverse partitions, additional qualitative results, comparisons with baselines, and failure cases. Our code is in \texttt{code/}. 
Code, visualizations of the forward and reverse partitions, additional qualitative results, comparisons with baselines, and failure cases are available at \url{https://atlas-track-objects.github.io/}.

\subsection{Defining transformations}
\label{define_transformations}
A relevant question that readers might have is what we define as transformations. We frame this definition by looking to cognitive science. Classical models of object perception in cognitive science~\cite{spelke1992} identify objects as being \emph{cohesive}; i.e., the points on them form a connected surface at all times. We therefore consider transformations as events that \emph{break} this property: either by causing sets of points to not be cohesive anymore (decomposition events) or by causing new sets of points to begin to move cohesively (assembly events).

\subsection{Dataset Details}
\label{dataset_details}

\begin{figure*}
    \centering
    \includegraphics[width=1\linewidth]{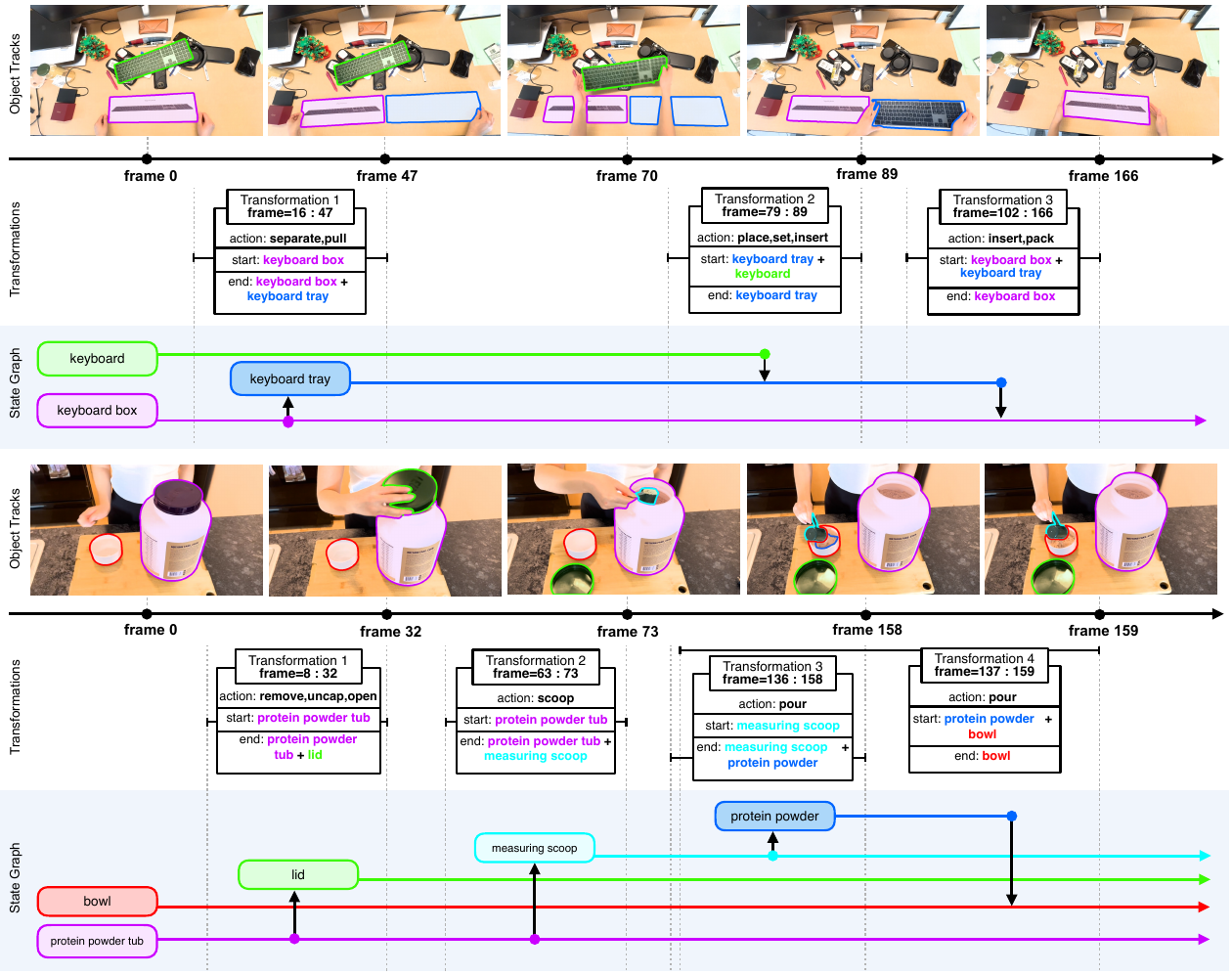}
    \caption{Examples of \dataset. Annotated object tracks and transformations allow us to construct a state graph. }
    \label{fig:dataset}
\end{figure*}

\textbf{Data Collection.} We collect videos spanning household activities, cooking, and assembly tasks captured from both egocentric and third-person viewpoints with no jump cuts. The resulting dataset includes a broad range of actions such as removal, insertion, opening, closing, cutting, assembly, breaking, attachment, unscrewing, capping, pouring, scooping, peeling, connecting, splitting, and tearing. These actions are applied across diverse objects ranging from food items (e.g., oranges, pineapples, broccoli, peppers, eggs, pears) to household items (e.g., bottles, blenders, jars, sunglasses, toothbrushes, instruments), electronics (e.g., phones, earbuds, remotes, calculators, keyboards), furniture parts (e.g., chairs), and toys (e.g., LEGO bricks, train tracks, Play-Doh). Videos are sourced from: (1) recordings by the authors with an iPhone; (2) YouTube tutorials retrieved via action-object queries (e.g., ``how to replace Swiffer mop pad'', ``how to mount a car tire''); and (3) existing benchmarks including VOST~\cite{vost}, VSCOS~\cite{vscos}, M$^3$-VOS~\cite{m3vos}, and IKEA Video Manuals~\cite{liu2024ikea}.

\textbf{Annotation Protocol.} We annotate decomposition and assembly events only when all participating objects can be clearly identified before and after the transformation, so that every annotated event has pre- and post-transformation objects.

\textbf{Annotation Pipeline.} We first identify all objects participating in transformations and provide per-frame segmentation masks throughout the video. Given the density of per-frame annotation, we provide sparse manual masks in our first pass on key frames, and use SAM2~\cite{sam2} to automatically propagate the masks across the sequence. Then, we review the masks by SAM2 and correct them where necessary using a polygon-add and polygon-erase tool to finalize segmentation masks for each object $o_j$. Finally, we annotate each transformation $\tau_i$ with its temporal boundaries ($t_i^s, t_i^e$), a free-text action description ($v_i$), and the pre- and post- object instances ($\mathcal{O}_i^{\text{pre}}$, $\mathcal{O}_i^{\text{post}}$).

\textbf{Dataset Statistics.}
\dataset contains $97$ videos with an average length of $13.6$ seconds, annotated at $15$ fps for a total of $19{,}739$ frames. It has $352$ objects ($3.6$ per video on average) across $54{,}369$ masks, and $302$ transformations ($144$ assemblies and $158$ decompositions, $3.1$ per video on average). Of the $97$ videos, $27$ contain only decomposition events, $23$ contain only assembly events, and $47$ contain both transformation types. Qualitative examples are provided in Figure~\ref{fig:dataset}.

\subsection{Evaluation on Prompt-Based Benchmarks}
\label{prompt_based_eval}

\textbf{Incorporating Prompt Masks.}
VOST~\cite{vost}, VSCOS~\cite{vscos}, M$^3$-VOS~\cite{m3vos}, and VOST-TAS~\cite{tubeletgraph} provide a first-frame mask for each prompt object.
For comparison with prompt-based methods on these benchmarks, we incorporate the prompt masks into the initial entity set of the forward partition.
Following TubeletGraph~\cite{tubeletgraph}, we first resolve overlaps between the prompt masks and CropFormer~\cite{cropformer} entities.
For each CropFormer entity, we compute the fraction of its area overlapping with the union of prompt masks.
We keep entities with overlap at most $\tau_{\text{coverage}} = 0.25$ unchanged, remove overlapping pixels for overlap between $\tau_{\text{coverage}}$ and $\tau_{\text{remove}} = 0.75$, and discard entities with overlap at least $\tau_{\text{remove}}$.
We then initialize a separate tubelet for each prompt mask, retaining its correspondence to the original prompt for evaluation.
The reverse partition is constructed from the last frame as usual.

\textbf{Object Tracking Evaluation.}
For VOST~\cite{vost}, VSCOS~\cite{vscos}, and M$^3$-VOS~\cite{m3vos}, we derive the predicted masks for each prompt object as follows. Initially, the prompt object is represented by the tubelet initialized from its first-frame mask. Following a detected decomposition, we track the resulting objects and output the union of their masks. For assembly, if the prompt object merges into another object, its predicted track terminates at the detected transformation. If another object merges into the prompt object, we continue tracking the resulting object, whose masks include both objects.

For M$^3$-VOS, we exclude $20$ videos that continue annotating the inserted object after assembly (e.g., tracking a puzzle piece separately after it has been inserted into a puzzle), whereas our formulation  updates the larger object's mask to include the inserted object and no longer tracks the inserted object separately. All methods are evaluated on the same remaining videos.

\textbf{Transformation Detection Evaluation.}
VOST-TAS~\cite{tubeletgraph} annotates only transformations involving the prompt object.
We therefore evaluate only predicted transformations involving the prompt object's track or its transformation products in the predicted state graph, discarding all other predictions.

\subsection{Additional Results}
\label{additional_results}
Tables~\ref{tab:assembly_results} and~\ref{tab:decomp_results} provide a breakdown of transformation detection performance by transformation type. Table~\ref{tab:bootstrap} reports a paired bootstrap comparison between
\ours and each baseline over the 97 videos of \dataset, using 10{,}000
resamples, confirming that the improvements in Table~\ref{tab:main} are statistically significant.

\begin{table*}[t]
\centering
\setlength{\tabcolsep}{4pt}
\renewcommand{\arraystretch}{0.95}

\begin{tabular*}{\textwidth}{@{\extracolsep{\fill}}l cc cc cc cc@{}}
\toprule

\textbf{Method}
& \multicolumn{2}{c}{Temp. Loc.}
& \multicolumn{2}{c}{Sem. Acc.}
& \multicolumn{2}{c}{Spatiotemp.}
& \multicolumn{2}{c}{Overall} \\

\cmidrule(lr){2-3}
\cmidrule(lr){4-5}
\cmidrule(lr){6-7}
\cmidrule(lr){8-9}

& $\mathcal{T}_P$ & $\mathcal{T}_R$
& $\mathcal{S}_V$ & $\mathcal{S}_O$
& $\mathcal{H}_{ST,P}$ & $\mathcal{H}_{ST}$
& $\mathcal{H}_P$ & $\mathcal{H}$ \\

\midrule

GPT-4.1 + GDINO
& \textbf{92.0} & \textbf{72.2}
& 82.7 & 95.2
& 12.4 & 9.7
& 8.8 & 6.9 \\

GPT-4.1 + SAM3
& \textbf{92.0} & \textbf{72.2}
& 82.7 & 94.4
& 21.2 & 16.7
& 10.6 & 8.3 \\

\midrule

TubeletGraph
& -- & --
& -- & --
& -- & --
& -- & -- \\

FluxGraph
& -- & --
& -- & --
& -- & --
& -- & -- \\

TubeletGraph-R
& 30.2 & 39.6
& 52.6 & 64.0
& 13.2 & 17.4
& 5.8 & 7.6 \\

TubeletGraph-Bidir
& 30.2 & 39.6
& 52.6 & 67.9
& 13.8 & 18.1
& 6.9 & 9.0 \\

\quad + Verif. \& Realign.
& 66.0 & 44.4
& 81.2 & 95.4
& 29.9 & 20.1
& 25.8 & 17.4 \\

\midrule

\textbf{\ours}
& 77.6 & 52.8
& \textbf{89.5} & \textbf{97.1}
& \textbf{59.2} & \textbf{40.3}
& \textbf{51.0} & \textbf{34.7} \\

\bottomrule
\end{tabular*}

\caption{Comparison with baselines on \dataset for assembly transformations. TubeletGraph and FluxGraph only track forward and cannot detect assembly.}
\label{tab:assembly_results}

\end{table*}

\begin{table*}[t]
\centering
\setlength{\tabcolsep}{4pt}
\renewcommand{\arraystretch}{0.95}

\begin{tabular*}{\textwidth}{@{\extracolsep{\fill}}l cc cc cc cc@{}}
\toprule

\textbf{Method}
& \multicolumn{2}{c}{Temp. Loc.}
& \multicolumn{2}{c}{Sem. Acc.}
& \multicolumn{2}{c}{Spatiotemp.}
& \multicolumn{2}{c}{Overall} \\

\cmidrule(lr){2-3}
\cmidrule(lr){4-5}
\cmidrule(lr){6-7}
\cmidrule(lr){8-9}

& $\mathcal{T}_P$ & $\mathcal{T}_R$
& $\mathcal{S}_V$ & $\mathcal{S}_O$
& $\mathcal{H}_{ST,P}$ & $\mathcal{H}_{ST}$
& $\mathcal{H}_P$ & $\mathcal{H}$ \\

\midrule

GPT-4.1 + GDINO
& 69.9 & 58.9
& \textbf{98.9} & 96.7
& 7.5 & 6.3
& 6.8 & 5.7 \\

GPT-4.1 + SAM3
& 69.9 & 58.9
& \textbf{98.9} & \textbf{98.6}
& 17.3 & 14.6
& 16.5 & 13.9 \\

% Gemini 3.1 Pro + SAM3 
% & \textbf{95.8} & \textbf{86.1} 
% & 97.8 & 96.0 & 23.2 & 20.9 & 21.1 & 19.0 \\

\midrule

TubeletGraph
& 37.7 & 50.6
& 80.0 & 91.0
& 22.6 & 30.4
& 16.5 & 22.2 \\

FluxGraph
& 39.4 & 31.6
& 82.0 & 88.9
& 23.6 & 19.0
& 16.5 & 13.3 \\

TubeletGraph-R
& -- & --
& -- & --
& -- & --
& -- & -- \\

TubeletGraph-Bidir
& 37.7 & 50.6
& 78.8 & 91.7
& 22.6 & 30.4
& 17.0 & 22.8 \\

\quad + Verif. \& Realign.
& 62.7 & 56.3
& 92.1 & 95.8
& 38.7 & 34.8
& 31.7 & 28.5 \\

\midrule

\textbf{\ours}
& \textbf{77.3} & \textbf{62.7}
& 90.9 & 96.7
& \textbf{55.5} & \textbf{44.9}
& \textbf{46.1} & \textbf{37.3} \\

\bottomrule
\end{tabular*}

\caption{Comparison with baselines on \dataset for decomposition transformations. TubeletGraph-R only tracks in reverse and cannot detect decomposition.}
\label{tab:decomp_results}

\end{table*}
 
\newcommand{\diffci}[3]{\(\mathbf{#1}\,(#2, #3)\)}

\begin{table*}[t]
\centering
\small
\setlength{\tabcolsep}{2pt}
\begin{tabular*}{\textwidth}{@{\extracolsep{\fill}}lccccc@{}}
\toprule
\textbf{\ours\ $-$ Method}
& $\Delta\gJ_{\text{tubelet}}$
& $\Delta\mathcal{H}_{ST,P}$
& $\Delta\mathcal{H}_{ST}$
& $\Delta\mathcal{H}_P$
& $\Delta\mathcal{H}$ \\
\midrule
GPT-4.1 + GDINO
& \diffci{16.2}{10.7}{21.6}
& \diffci{47.3}{38.2}{56.3}
& \diffci{34.8}{26.6}{42.6}
& \diffci{40.5}{31.2}{50.0}
& \diffci{29.8}{21.9}{37.8} \\
GPT-4.1 + SAM3
& \diffci{15.0}{9.2}{20.9}
& \diffci{38.0}{27.8}{47.9}
& \diffci{27.2}{17.8}{35.9}
& \diffci{34.4}{24.7}{44.2}
& \diffci{24.8}{16.5}{33.2} \\
\midrule
TubeletGraph
& \diffci{11.4}{7.1}{15.5}
& \diffci{34.4}{25.0}{44.2}
& \diffci{26.8}{19.7}{34.2}
& \diffci{31.7}{22.3}{41.0}
& \diffci{24.5}{17.0}{32.1} \\
FluxGraph 
& \diffci{20.7}{15.9}{25.4}
& \diffci{33.5}{21.8}{45.1}
& \diffci{32.8}{25.4}{40.3}
& \diffci{31.7}{20.9}{42.0}
& \diffci{29.1}{21.6}{36.7} \\
TubeletGraph-R
& \diffci{15.9}{12.4}{19.4}
& \diffci{43.9}{35.2}{52.9}
& \diffci{34.4}{27.9}{41.1}
& \diffci{42.4}{33.5}{51.6}
& \diffci{32.5}{25.1}{39.9} \\
TubeletGraph-Bidir
& \diffci{12.6}{7.8}{17.7}
& \diffci{38.6}{30.9}{46.8}
& \diffci{18.2}{11.3}{25.3}
& \diffci{36.0}{27.8}{44.5}
& \diffci{19.9}{12.3}{27.4} \\
\quad + Verif.\ \& Realign.
& \diffci{12.7}{7.9}{17.7}
& \diffci{21.9}{15.4}{28.8}
& \diffci{14.9}{9.3}{20.8}
& \diffci{18.9}{12.2}{25.6}
& \diffci{12.9}{7.6}{18.3} \\

\bottomrule
\end{tabular*}
\caption{Paired bootstrap differences (\ours{} minus baseline) on
\dataset, with 95\% confidence intervals in parentheses.}\label{tab:bootstrap}
\end{table*}

\subsection{System Analysis}
\label{robustness}
\textbf{Hyperparameters.} We set $\tau_{\text{coverage}}=0.25$ as in TubeletGraph, and fix
$\tau_{\text{dup}}=0.9$ for constructing the forward and reverse
spatiotemporal partitions, $\omega_{\text{long}}=4\,\mathrm{s}$ and
$\omega_{\text{short}}=1\,\mathrm{s}$ for verification and realignment,
and $\tau_{\text{merge}}=0.75$ and $\tau_{\text{area}}=0.2$ for
finalizing the unified pool of tubelets, all without tuning.
For
detecting state changes, we select $\tau_{\text{contain}}=0.7$,
$\tau_{\text{IoU}}=0.05$, and $\omega=1\,\mathrm{s}$ on a separate set
of videos recorded by the authors, and use them
for all datasets without further tuning.

\textbf{Robustness.} Table~\ref{tab:sensitivity} varies $\tau_{\text{contain}}$,
$\tau_{\text{IoU}}$ and $\omega$, which are used for transformation detection. \ours is insensitive to $\omega$, with $\mathcal{H}$ changing by at most $0.7$ points. \ours is also insensitive to $\tau_{\text{contain}}$ for $\tau_{\text{IoU}}$ up to $0.1$: across that range $\mathcal{H}$ stays within $2.0$ points of the default and exceeds TubeletGraph-Bidir with the same VLM stage ($\mathcal{H} = 23.2$) by at least $11.2$ points. Intuitively, $\tau_{\text{IoU}}$ bounds the size of the matched tubelet relative to the emerging one, preventing a new object from being matched to a much larger tubelet that only coincidentally covers it, and the VLM does not always reject such candidates. Raising $\tau_{\text{IoU}}$ from $0$ to $0.05$ thus improves $\mathcal{H}_P$ at every $\tau_{\text{contain}}$. However, at $\tau_{\text{IoU}} = 0.2$, it discards many transformations in which a small object separates from or merges into a larger one. $\mathcal{H}$ drops substantially, but $\mathcal{H}_P$ drops too because the discarded candidates are now predominantly correct while false positives remain.

\textbf{Computational Efficiency.}
The main computational bottleneck of \ours is constructing the forward and reverse spatiotemporal partitions. On \dataset, entity segmentation takes $1.25$ seconds per frame while tracking takes $2.97$ seconds per frame on a single NVIDIA RTX A6000 GPU, as \ours propagates all entities jointly in a single SAM2~\cite{sam2} pass per direction. TubeletGraph~\cite{tubeletgraph} also segments every frame, but tracking in the forward direction takes $13.04$ seconds per frame, as it tracks one prompted entity at a time, requiring a separate pass for each entity detected by CropFormer~\cite{cropformer} in the first frame. Tracking in reverse as well, as in TubeletGraph-Bidir, doubles this tracking cost. FluxGraph~\cite{fluxgraph}, an efficient variant of TubeletGraph, reduces entity segmentation to $0.47$ seconds per frame, but it also tracks one entity at a time, totaling $6.45$ seconds per frame.

\begin{table}[t]
\centering
\small
\setlength{\tabcolsep}{4.5pt}
\begin{tabular}{@{}l cc cc cc@{}}
\toprule
& \multicolumn{2}{c}{$\tau_{\text{contain}}=0.6$}
& \multicolumn{2}{c}{$\tau_{\text{contain}}=0.7$}
& \multicolumn{2}{c}{$\tau_{\text{contain}}=0.8$} \\
\cmidrule(lr){2-3}\cmidrule(lr){4-5}\cmidrule(lr){6-7}
& $\mathcal{H}_P$ & $\mathcal{H}$
& $\mathcal{H}_P$ & $\mathcal{H}$
& $\mathcal{H}_P$ & $\mathcal{H}$ \\
\midrule
$\tau_{\text{IoU}}=0$     & 39.1 & 35.8 & 43.1 & 38.1 & 44.2 & 36.4 \\
$\tau_{\text{IoU}}=0.025$ & 46.0 & 36.4 & 46.8 & 36.8 & 47.3 & 35.1 \\
$\tau_{\text{IoU}}=0.05$  & 48.5 & 36.8 & \cellcolor{gray!20}48.2 & \cellcolor{gray!20}36.1 & 52.8 & 37.4 \\
$\tau_{\text{IoU}}=0.1$   & 52.5 & 35.1 & 53.8 & 35.4 & 53.3 & 34.4 \\
$\tau_{\text{IoU}}=0.2$   & 45.3 & 22.2 & 49.7 & 23.8 & 48.9 & 22.5 \\
\midrule
& \multicolumn{2}{c}{$\omega=0.5$\,s}
& \multicolumn{2}{c}{$\omega=1$\,s}
& \multicolumn{2}{c}{$\omega=2$\,s} \\
\cmidrule(lr){2-3}\cmidrule(lr){4-5}\cmidrule(lr){6-7}
& $\mathcal{H}_P$ & $\mathcal{H}$
& $\mathcal{H}_P$ & $\mathcal{H}$
& $\mathcal{H}_P$ & $\mathcal{H}$ \\
\midrule
& 50.0 & 36.1 & \cellcolor{gray!20}48.2 & \cellcolor{gray!20}36.1 & 47.0 & 36.8 \\
\bottomrule
\end{tabular}
\caption{We vary $\tau_{\text{contain}}$ and $\tau_{\text{IoU}}$ used for transformation detection with $\omega=1$\,s (top), and the temporal search window $\omega$ with $\tau_{\text{contain}}=0.7$ and $\tau_{\text{IoU}}=0.05$ (bottom). Shaded cells indicate the default configuration.}
\label{tab:sensitivity}
\end{table}

\subsection{Limitations}

\ours detects a transformation only when a tubelet newly emerges in the
forward or reverse direction. Transformations that do not split or
merge objects, such as gradual in-place state changes (e.g., melting or
deformation), produce no new tubelets and are not detected. Moreover,
like TubeletGraph~\cite{tubeletgraph}, \ours relies on the tracker's
false negatives: the tracker must reliably fail to extend an existing mask over a transformation product for a new tubelet to emerge.  Finally, constructing both a forward and a
reverse spatiotemporal partition is costly. However, the two partitions are fully independent and can be computed in parallel, and improvements in efficiency~\cite{fluxgraph} directly reduce the cost of both.

\end{document}